\documentclass[conference]{IEEEtran}
\IEEEoverridecommandlockouts
\usepackage{cite}
\usepackage{amsmath,amssymb,amsfonts}
\usepackage{algorithmic}
\usepackage{graphicx}
\usepackage{textcomp}
\usepackage{xcolor}
\def\BibTeX{{\rm B\kern-.05em{\sc i\kern-.025em b}\kern-.08em
    T\kern-.1667em\lower.7ex\hbox{E}\kern-.125emX}}
\usepackage{mathtools}
\usepackage{multirow}
\usepackage{float}
\usepackage{siunitx}
\usepackage{listings}
\usepackage{svg}
\usepackage[linesnumbered,ruled,vlined]{algorithm2e}
\usepackage{hyperref}

\begin{document}
\title{Hardware architecture for high throughput event visual data filtering with matrix of IIR filters algorithm\\
\thanks{The work presented in this paper was supported by the National Science Centre project no. 2021/41/N/ST6/03915 entitled ``Acceleration of processing event-based visual data with the use of heterogeneous, reprogrammable computing devices'' (first author) and partly supported by the programme ``Excellence initiative – research university'' for the AGH University of Science and Technology.}
}

\author{\IEEEauthorblockN{Marcin Kowalczyk}
\IEEEauthorblockA{\textit{Department of Automatic Control and Robotics} \\
\textit{AGH University of Science and Technology}\\
Krakow, Poland \\
kowalczyk@agh.edu.pl}
\and
\IEEEauthorblockN{Tomasz Kryjak}
\textit{Senior Member IEEE}\\
\IEEEauthorblockA{\textit{Department of Automatic Control and Robotics} \\
\textit{AGH University of Science and Technology}\\
Krakow, Poland \\
tomasz.kryjak@agh.edu.pl}
}


\maketitle


\begin{abstract}

Neuromorphic vision is a rapidly growing field with numerous applications in the perception systems of autonomous vehicles. 
Unfortunately, due to the sensors working principle, there is a significant amount of noise in the event stream.
In this paper we present a novel algorithm based on an IIR filter matrix for filtering this type of noise and a hardware architecture that allows its acceleration using an SoC FPGA.
Our method has a very good filtering efficiency for uncorrelated noise -- over 99\% of noisy events are removed.
It has been tested for several event data sets with added random noise.
We designed the hardware architecture in such a way as to reduce the utilisation of the FPGA's internal BRAM resources.
This enabled a very low latency and a throughput of up to 385.8 MEPS million events per second. 
The proposed hardware architecture was verified in simulation and in hardware on the \textit{Xilinx Zynq Ultrascale+ MPSoC} chip on the \textit{Mercury+ XU9} module with the \textit{Mercury+ ST1} base board.


\end{abstract}


\section{Introduction}


Event cameras, also called dynamic or neuromorphic vision sensors, have become increasingly popular in recent years. 
Unlike ``traditional'' cameras, they do not transmit an image every fixed period of time in the form of a matrix containing information about the brightness level measured by successive pixels of the video sensor array. 
Instead, each pixel acts as an independent brightness sensor. 
When the brightness recorded by a given pixel changes, an event is sent from the sensor containing information about the timestamp, polarity (whether the brightness has increased or decreased), and the coordinates of the pixel for which the change occurred. 
Each pixel registers events independently, but they are all connected to a common clock based on which the timestamp is determined.


Event cameras are characterised by high dynamic range, high temporal resolution, and low power consumption. 
Compared to traditional video cameras, they perform better in adverse lighting conditions, record even very fast motion, and have low latency between an event (e.g. motion) and its transmission.


These sensors are rapidly gaining popularity and are constantly being developed. 
This is evidenced, among other things, by increasing resolution and increasing throughput (understood as the maximum number of events per unit of time).
In 2017, the best event sensor offered a resolution of \(640 \times 480\) and transmitted a maximum of 300 million events per second (MEPS). 
On the contrary, in 2022, the best sensors offer a resolution of \(1280 \times 720\) and a throughput of 1200 MEPS\cite{gallego2020event}. 
One event can have a different size (number of bits) depending on the resolution and the assumed number of bits per timestamp. 
Assuming that one event is 8 bytes, for a bit rate equal to 1 MEPS, the system must process \(8 \si{MB}\) of data per second. 
This means that up to 8 GB of data per second would be transmitted for a steady stream of 1000 MEPS.
For comparison, a classical video data stream with a resolution of \(3840 \times 2160\) and 60 frames per second (4K, Ultra High Definition) has a bit rate of approximately 1.5 GB/s. 
However, a situation where the event sensor will record the maximum number of events is very unlikely, and the event flow rate heavily depends on the dynamics of the observed scene.


The presence of noise in the data recorded from the sensors has a very large impact on the performance of the algorithms, as well as the entire system.
For example, it can significantly reduce the effectiveness of classification.
For this reason, data filtering is used in virtually all systems where sensor data is used, e.g. autonomous vehicles, advanced driver assistance systems (ADAS), safety systems and advanced video surveillance systems (AVSS), etc. 
In the case of currently available classical vision sensors, noise is usually in the form of additive Gaussian, i.e. a slight deviation of the measurement from its actual value, which makes it not so visible and has a relatively small impact on the performance of the system (with some exception in the case of low-light registration). 
The situation is different in the case of event sensors, where the disturbances have the form of impulse noise (i.e. events that do not actually occur in the observed scene). 
They are much more visible and can have a significant impact on the operation of a system with a DVS camera.


In this paper, we present a novel event data filtering algorithm and its hardware architecture. 
The algorithm has been programmed and tested in the \textit{MATLAB} computing environment. 
The hardware architecture was implemented and tested in a \textit{Xilinx Zynq Ultrascale+ MPSoC} device of \textit{Mercury+ XU9} module with \textit{Mercury+ ST1} base card. 
The designed architecture achieves a throughput of \(385.8 \si{MEPS}\) with low resource utilisation and low power consumption.


Therefore, the main contributions of this paper can be summarised as follows:
\begin{itemize}
    \item a novel event sensor data filtering algorithm that reduces memory usage, so it can be implemented in computing systems with small memory resources,
    \item a high-throughput hardware architecture that implements the proposed filtering algorithm,
    \item verification of the performance of the proposed architecture on the Xilinx Zynq Ultrascale+ MPSoC device of the Mercury+ XU9 module with the Mercury+ ST1 base card.
\end{itemize}
To the best of the authors' knowledge, the presented architecture provides the highest filtering throughput of video event sensor data among the solutions described in the scientific literature.


This paper has been divided into the following parts:
Section \ref{sec:previous} presents previously published research related to event sensor data filtering.
Then Section \ref{sec:algorytm} presents the proposed filtering algorithm.
The hardware architecture implementing the discussed algorithm is presented in Section \ref{sec:architektura}.
The evaluation of the proposed filtering method and the underlying architecture is described in Section \ref{sec:ewaluacja}.
The paper concludes with a summary and possible further development of the architecture described in Section \ref{sec:podsumowanie}.

\section{Previous work}
\label{sec:previous}


Due to the high practical importance of event camera data filtering, as described in the introduction, a number of papers on this topic have been published in the scientific literature.
Their analysis and comparison are presented in the following section.


One of the first filtering methods was proposed in the paper \cite{delbruck2008frame}. 
This filter only passes events for which there has been previous activity in their neighbourhood. 
This results in filtering out uncorrelated, isolated events. 
The filter uses a timestamp map of size equal to the sensor resolution multiplied by 2. 
An array of size \(128 \times 128 \times 2\), where each element has 32 bits, was used in the described case.
The filter parameter is the time interval for which a neighbourhood will be considered active after an event has been recorded. 
The processing time for a single event was \(0.1 \si{\micro s}\). 
This translates into a throughput of \(10 \si{MEPS}\). 
A laptop with a Pentium M processor clocked at \(2.13 \si{GHz}\) was used for processing. 
In the paper, a method for feature extraction and object tracking was also proposed.


In \cite{linares2015usb3}, a solution for filtering data from an event-based camera was described.
The authors proposed an FPGA-based framework for uncorrelated noise removal.
The solution is based on storing the timestamp corresponding to the processed event in a register array. 
For each input event, the timestamp is read, and then the difference between the timestamp of the processed event and the read one is calculated. 
If it is less than the set threshold, the event is forwarded. 
The event marker is written to the same memory location. 
It can also be written into neighbouring pixels to increase spatial correlation. 
In this paper, a timestamp matrix of size \(128 \times 128\) was used.
It supports the processing of sensor data with the same resolution or higher if the event address space is subsampled.
In the solution, a hardware platform based on a Lattice FPGA was used. 
The packets were transmitted at rates of \(10 \si{MEPS}\).
The authors also proposed an architecture for object tracking.


The authors of the paper \cite{czech2016evaluating} compared several methods for filtering event sensor data. 
The first three filters check the distance in time of the processed event with the last event in the neighbourhood. 
The first two methods have a different neighbourhood shape (8- or 4-element), while the third one also takes into account the event polarity. 
The fourth one requires that there are at least two events in the neighbourhood in a given time interval. 
The fifth filter removes data whose polarity is the same as the previous event reported by that pixel. 
The sixth and seventh filters only pass events if sufficient time has passed since the previous event generated by that pixel. 
However, the seventh takes polarisation into account. 
The eighth filter calculates the average time between events for each pixel. 
If this time is too short, the events are treated as noise. 
The computational efficiency for the proposed methods is not explicitly presented in this paper. 
Based on the processing time given for the test sets, it can be calculated that the average throughput was \(5.93 \si{MEPS}\).


In paper \cite{barrios2018less}, the LDSI (Less Data Same Information) filtering algorithm is proposed. 
It reduces the amount of data needed to be processed without reducing the amount of relevant information. 
The proposed solution can be configured to filter out fewer or more data. 
It was implemented on a \textit{Xilinx Virtex 6} FPGA chip to process data from a \(128 \times 128\) resolution sensor.
The filtering algorithm was based on spiking cells inspired by the action of biological neurones. 
Processing occurs in four serially connected layers. 
Each processed event increases the potential of the corresponding neurones in the layers.
The potential of neighbouring neurones also increases in the synaptic terminals (Alayer).
According to the information provided by the authors, the proposed architecture can operate at \(177 \si{MHz}\). 
However, it is not explicitly stated in the paper that there can be one event given to the input of the architecture for each clock cycle.


The filtering algorithm and its implementation in the \textit{IBM TrueNorth Neurosynaptic System} neuromorphic processor are presented in the paper \cite{padala2018noise}.
A spiking neural network was used for filtering, in which a variation of the \textit{integrate-and-fire} neurone model was applied. 
The proposed filter has two layers. 
The first introduces a refractive period, which limits the maximum response frequency of the neurones, thus removing high-frequency interference from a single pixel. 
The second layer is a neural implementation of the nearest-neighbour method. 
It checks whether other events were generated in the vicinity of the processed event by summing up synaptic activity. 
Note that due to the specification of the \textit{TrueNorth} processor, it was necessary to reduce the temporal resolution of the events to \(1 \si{ms}\). 
An event sensor with a resolution of \(304 \times 240\) was used. 
The throughput of the proposed solution is not given in the paper.


The authors of the paper \cite{bisulco2020near} proposed a method for filtering and compressing the event stream, which is based on two switched time windows.
The windows alternate in two phases. 
In the first phase, events are written into one window (creating an image representing the sensor data), while data from the second window are processed, and the window is cleared.
In the second phase, the first window is processed, and events are written into the second window.
For every two neighbouring pixels, a logical product is performed. 
If there are events with the same polarity in the pixels next to each other, they are sent to the aggregation module. 
This operation is performed separately for vertical and horizontal pixels. 
The aggregation module combines data from successive time slots until the number of events exceeds a predefined threshold. 
If this does not happen within five time windows, the buffer is cleared. 
The timestamp of individual events is not used and is not passed on to the filter output. 
The implementation was done for the \textit{Spartan-6} FPGA chip. 
A maximum operating frequency of \(51.18 \si{MHz}\) was achieved. 
The maximum event stream throughput is not explicitly stated in this work either. 
However, it is written that the average event rate at the output of the sensor used was \(50 \si{MEPS}\).


In the paper \cite{xiao2021snn}, a spiking neural network with adaptive time window length is proposed. 
It uses the leaky integrate-and-fire (LIF) neurone model. 
The authors claim that the proposed method performs better than classical methods using time-window filtering. 
No information was found about the throughput of the proposed solution.


In paper \cite{mohamed2022dba}, an adaptive background activity filtering method is proposed for event data. 
It uses the KNN (K-nearest neighbour) non-parametric regression algorithm and optical flow.
The proposed method consists of two steps. The objective of the first one is to discard the supermirror events using dynamic timestamping techniques. The goal of the second stage, in turn, is to remove background noise using an adaptive KNN algorithm.
The authors claim that this method achieves a high SNR (signal to noise ratio) parameter, at \(13.64 \si{dB}\).
The system was run on a \textit{Jetson TX2} card and operated in real time for a sensor with a resolution of 240 x 180. The authors claim that the sensor used can generate up to 3 million events per second. However, they did not provide information on whether the proposed solution could be used for higher throughput.


The literature review performed confirms the importance of event data filtering. Methods to remove distortions at the DVS output are constantly being developed. Newer and newer algorithms are proposed that allow for increasingly efficient data processing. The highest throughput among the analysed articles had the method proposed in \cite{bisulco2020near}, allowing processing 50 million events per second. Architecture presented in this work can process up to 385.8 million events per second.

\section{The Proposed Filtering Algorithm}
\label{sec:algorytm}


The filter presented was designed for its hardware implementation using a fine-grained pipelined architecture so as to maximise the bandwidth of the architecture while maintaining high filtering efficiency. 
As such, particular attention has been paid to reducing memory usage. 
This is dictated by the fact that the resolution of the event sensors is constantly increasing. 
For HD resolution (\(1280 \times 720\)), an algorithm storing the last timestamp for each pixel would require approximately \(1280 \cdot 720 \cdot 32 \approx 29.49 \si{Mb}\) of memory.
It should be noted here that the SoC FPGA chip used in this project contains only \(11 \si{Mb}\) of BRAM memory. 
An additional complication would be to read data from several memory cells (those that lie in the vicinity of the event being processed). 
This would require a reduction in the throughput of the algorithm-based architecture or a significant increase in the use of memory resources (linear to the number of cells from which a read is required in each clock).


To reduce memory requirements, it was decided to divide all pixel coordinates into square areas.
Only one timestamp was stored for each area. 
This reduced the memory requirement as a function of the square of the scale factor (side length of the square in pixels). 
For areas of size \(20 \times 20\), the memory requirement for a sensor with HD resolution is approximately \(1280/20 \cdot 720/20 \cdot 32 \approx 73.73 \si{Kb}\).


However, for an area-based method, it is not sufficient to simply remember the timestamp value of the last event that occurred in the area.
Such an approach would result in very mild filtering that would remove only a small portion of the noise. 
This is because increasing the size of the areas increases the probability that more than one disturbance will be assigned to one area in a small interval. 
Instead, it was decided to use an independent IIR filter for each area. 
The proposed method is therefore based on a matrix of IIR filters whose outputs are the event cut-off times for each area. 
For each processed event, the filter state corresponding to the area to which it belongs is updated.


In the proposed approach, it proved to be a problem that a large number of correct events were removed when no event has been received in the area for a long time.
If a time equal to 2 seconds has passed since the last update of the considered area, at least 33 consecutive events will be removed. 
To minimise this effect, a global filter matrix update functionality was introduced to the algorithm. 
It consists in modifying the state of the filters every specified time or every specified number of processed events. 
However, only the filters for which there has been no change in state since the last global update are modified.
This mechanism is designed to weaken the filtering threshold for areas that have not been active for a long time. 

\begin{figure}[!t]
	\centering
	\includegraphics[width=3.45in]{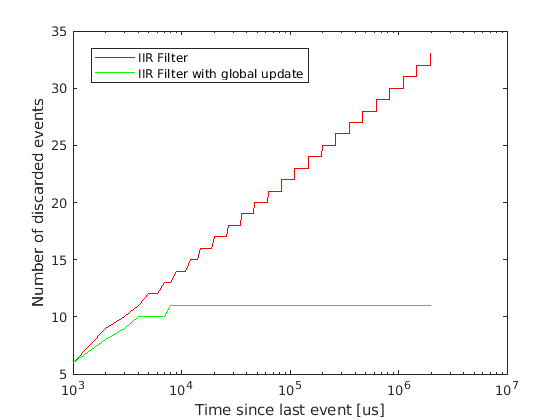}
	\caption{Number of discarded events in the function of time since the last filter update.}
	\label{fig:DiscardedEvents}
\end{figure}

Figure \ref{fig:DiscardedEvents} shows a comparison of the number of deleted events versus time without update for the algorithm with and without global filter update. 
These graphs were obtained for a filter length of \(200 \si{\micro s}\), an update factor of \(0.25\), and a global update was performed every \(1000 \si{\micro s}\). 
It can be seen that for the parameters used, the number of removed events saturates reaching a value of \(11\) for a time since the last event of \(8000 \si{\micro s}\). 
The Algorithm \ref{alg:event_filter} presents the pseudocode of the proposed filtering.

\begin{algorithm}[!t]
 \KwData{
    events - event vector\newline
    scale - size of a single area\newline
    sizeDVS - sensor resolution\newline
    filterLength - filtration time threshold\newline
    updateFactor - filter update rate\newline
    TimeMap - filter state matrix}
 \KwResult{result - vector of filtration results}
 \(index \gets 0\)\;
 \(ActiveMap \gets zeros(floor(sizeDVS/scale))\)\;
 \ForEach{\(event \in events\)}{
    \(xCell \gets floor(event.x/scale)\)\;
    \(yCell \gets floor(event.y/scale)\)\;
    \(thrTs \gets TimeMap[yCell, xCell]\)\;
    \(diffTs \gets event.ts - thrTs\)\;
    \(result[index] \gets diffTs < FilterLength\)\;
    \(index \gets index + 1\)\;
    \(TimeMap[yCell, xCell] \gets thrTs \cdot (1-updateFactor) + event.ts \cdot updateFactor\)\;
    \(ActiveMap[yCell, xCell] \gets 1\)\;
    \(currentTs \gets event.ts\)\;}
 \For{\(yCell \gets 0\) \textbf{to} \(floor(sizeDVS[0]/scale)\)} {
    \For{\(xCell \gets 0\) \textbf{to} \(floor(sizeDVS[1]/scale)\)} {
        \If{\(\neg ActiveMap[yCell, xCell]\)} {
            \(TimeMap[yCell, xCell] \gets thrTs \cdot (1-updateFactor) + currentTs \cdot updateFactor\)\;
        }
    }
 }
 \caption{Event filtration}
 \label{alg:event_filter}
\end{algorithm}

\begin{figure}[!t]
	\centering
	\includegraphics[width=3.45in]{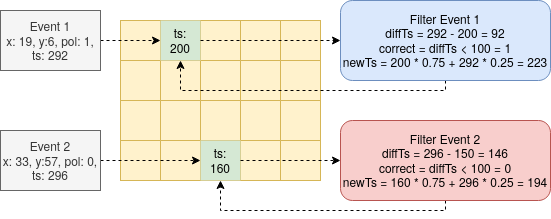}
	\caption{An example of how the discussed filtering algorithm works. For each event, the coordinates of the area to which it belongs are calculated. The timestamp difference is then calculated. If it is less than the filter length, the event is passed to the output. Finally, the timestamp in the matrix is updated.}
	\label{fig:AlgorytmFilter}
\end{figure}

Figure \ref{fig:AlgorytmFilter} shows an example of how the filtering works for two events and a sensor with a very low resolution of (\(80 \times 64\)). The size of the areas is \(16 \times 16\), the filter length is \(100 \si{us}\) and the update factor is \(0.25\). The areas to which the considered events belong are marked in green. The calculations for the first event are shown in the blue block. The calculated time lag difference is smaller than the filter length, so the event is valid and will be passed to the output. For the second event, the difference is smaller than the filter length, so it will be rejected. A new timestamp is then calculated for each event, which will be written into the corresponding area.

\begin{figure}[!t]
	\centering
	\includegraphics[width=3.45in]{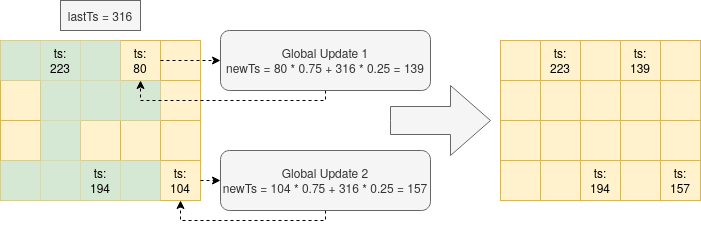}
	\caption{An example of how the global update works. The algorithm iterates through all areas. If the processed one is marked as not updated (yellow in the figure) it modifies its timestamp according to the given formula based on the timestamp of the last processed event.}
	\label{fig:AlgorytmUpdate}
\end{figure}

Figure \ref{fig:AlgorytmUpdate} shows an example of a global update. Areas where at least one event from the processed packet belonged are highlighted in green. These areas will be skipped. Above the matrix on the left was given the timestamp of the last event in the processed packet. The coefficients of the yellow areas will be updated according to the calculations shown. A timestamp matrix with example results is shown on the right.

\section{The Proposed Hardware Architecture}
\label{sec:architektura}

\begin{figure}[ht]
	\centering
	\include[width=3.45in]{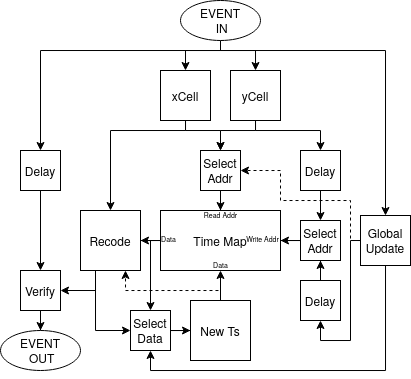}
	\caption{Diagram of the proposed hardware architecture.}
	\label{fig:architektura}
\end{figure}

Due to the increasing resolution and throughput of the event sensors, the hardware architecture implementing the filtering algorithm presented in Section \ref{sec:algorytm} was designed to maximise the number of events that can be processed in time unit. 
For this purpose, it was necessary to be able to process new events in consecutive clock cycles and to maximise the frequency for which it could operate. 
Its schematic is presented in Figure \ref{fig:architektura}.

The input data stream \texttt{EVENT IN} uses the \textit{AXI4-Stream} interface. 
It contains the signals \texttt{tvalid}, \texttt{tdata}, \texttt{tuser}, \texttt{tready}, and \texttt{tlast}.
The output data stream \textit{EVENT OUT} uses the same interface and an additional signal \texttt{correct}, which is the binary output of the filter. 
This signal has a value of \(0\) when the event is classified as noise.



The parameters of the designed architecture are:
\begin{itemize}
    \item \texttt{DATA\_WIDTH} -- the number of bits allocated to each event (and thus the width of the \texttt{tdata} signal),
    \item \texttt{SCALE} -- the width of the areas into which the sensor matrix has been divided,
    \item \texttt{SENSOR\_HEIGHT} -- sensor height,
    \item \texttt{SENSOR\_WIDTH} -- sensor width,
    \item \texttt{UPDATE\_FACTOR} -- filter state update factor,
    \item \texttt{FILTER\_LENGTH} -- filter length.
\end{itemize}


To simplify the architecture, the \texttt{SCALE} and \texttt{UPDATE\_FACTOR} parameters were assumed to be powers of 2. 
This allowed divisions (for \texttt{SCALE}) and multiplications (for \texttt{UPDATE\_FACTOR}) to be converted into bit shifts.
However, it is worth noting that modifying the architecture to support arbitrary values of these parameters is very simple. 
The first would require a divider in the \texttt{xCell} and \texttt{yCell} modules. 
The second would require two multipliers inside the \texttt{New Ts} module and possibly additional recoding levels inside the \texttt{Recode} module (due to the additional delay between reading data from memory and the corresponding write).
The architecture is designed so that data processing is stopped when the \texttt{tready} signal from the \texttt{EVENT OUT} interface is in the low state. 
This signal indicates that subsequent modules are unable to receive the data, so further processing could result in data loss.

\subsection*{xCell and yCell}
\label{xcell_ycell}


The modules \texttt{xCell} and \texttt{yCell} are responsible for determining the coordinates of the area of the currently processed event. 
On the input, they take its horizontal and vertical coordinates, respectively, and on the output, they give the coordinates of the area to which it belongs. 
To simplify the architecture, these modules perform a bit shift of the input event address. 
The output of both modules is then combined to create an address for the BRAM within \texttt{Time Map}. 
This address is also delayed inside the \texttt{Delay} as it is needed to store the updated filter state in the same BRAM memory cell.

\subsection*{Select Addr and Select Data}
\label{select_addr}

The modules \texttt{Select Addr} and \texttt{Select Data} control the addresses and data sent to \texttt{Time Map}. 
The inputs of \texttt{Select Addr} are the output addresses of \texttt{xCell} and \texttt{yCell} and the output address of \texttt{Global Update}.
The outputs are the read and write addresses to the BRAM storing the filter states.
The inputs of \texttt{Select Data} are data from \texttt{Time Map} and \texttt{Recode}.
The outputs are data to be written to the BRAM storing the filter states.
These modules also receive a flag indicating a global update process.
They are responsible for passing appropriate signals to the output, depending on whether data from the \texttt{EVENT IN} interface are being processed or the global update process is active.
The switching is controlled by the input flag.

\subsection*{Time Map}
\label{time_map}


The \texttt{Time Map} contains a BRAM that stores the filter state array. 
The inputs of the module are the read and write addresses from the corresponding \texttt{Select Addr} modules and the new filter state calculated inside the \texttt{New Ts}. 
The module output is the filter state corresponding to the read address, which is passed to \texttt{Recode}.
A simple dual-port BRAM with registered output was used. 
Consequently, reading the data takes two clock cycles. 
In addition to the signals given earlier, the memory requires two more: \texttt{Register B Clock Enable} and \texttt{Port A Write Enable}.
The first of these inputs receives the \texttt{tready} signal of the \texttt{EVENT OUT} interface. 
This causes the memory reading to be paused when data cannot be sent from the module. 
The signal given to the second input depends on the state of the flag that indicates a global update. 
If it is high, the negated value from the modified areas map (from \texttt{Global Update}) is given. 
Otherwise, the value of the delayed logical product of the \texttt{tvalid} and \texttt{tready} signals of the \texttt{EVENT IN} interface is given.

\subsection*{Recode}
\label{recode}


The \texttt{Recode} module is responsible for swapping the filter states read from the BRAM. 
The inputs of the module are the coordinates of the area from \texttt{xCell} and \texttt{yCell}, the output data of \texttt{Time Map}, and the output data of \texttt{New Ts}. 
A modified filter state is given at the output. 
The described element is needed because of the delay between receiving a new event and updating the filter state corresponding to this event. 
This is problematic when consecutive events belong to the same area. 
This is due to the fact that the states calculated in the previous clock cycles will not yet be taken into account in the BRAM of \texttt{Time Map}. 
Therefore, inside the module, it is necessary to check if the coordinates of the area of the new event are the same as any of the events from the previous three clock bars (the results are written into flags indicating data swapping). 
Two clock cycles result from the latency of the reading from BRAM memory. 
The third one, in turn, protects against collisions of reads and writes to the memory. 
A collision is a situation where the read and write addresses are the same. 
Very often, this results in reading incorrect data. 
If any of the flags mentioned is high, the data received from \texttt{New Ts} are passed to the output (must be appropriately delayed).

\subsection*{Verify}
\label{verify}


The module \texttt{Verify} checks whether the event should be filtered out. 
The inputs of the module are the delayed event from \texttt{EVENT IN} and the filter state from \texttt{Recode}.
The output is the \textit{AXI4-Stream} \texttt{EVENT OUT} interface, which contains the \texttt{correct} flag in addition to the basic signals. 
The filter length (architecture parameter) is added to the filter state. 
The result is compared with the timestamp of the received event. 
If it is greater, the data are classified as noise, and the flag is set to 0. 
Otherwise, it is set to 1.

\subsection*{Global update}
\label{global_update}

The module \texttt{Global Update} is responsible for the global update functionality. 
The input of the module is an event stream. 
The outputs are addresses of subsequent areas passed to \texttt{Time Map} and a flag indicating that the process is in progress passed to \texttt{Select Addr} and \texttt{Select Data}. 
For each input event, the current timestamp value needed for the update process is stored. 
Areas that received at least one event are also marked in the update matrix (and are skipped in the process). 
The process itself starts when the \texttt{tlast}, \texttt{tvalid}, and \texttt{tready} signals of the \texttt{EVENT IN} interface are high. 
This indicates the last event of the transmitted packet. 
The flag indicating blocking of the \texttt{EVENT IN} interface is then set to high state (the \texttt{tready} signal is set to 0).
The module then waits until the processing of the received events is completed (3 clock cycles, provided that the \texttt{tready} signal of the \texttt{EVENT OUT} interface is equal to 1). 
After this time, a flag is set indicating the update execution status, and the generation of all area addresses begins. 
The flags in the update matrix are also read. 
These are then delayed and passed to the \texttt{Write Enable} input of the BRAM. 
When the last one is passed to the output, the flag mentioned is set to low. 
Then, the system waits three clock cycles, until the new data are written to the BRAM of \texttt{Time Map}.
The input \texttt{EVENT IN} interface is unlocked (by setting the \texttt{tready} signal to high state), and the designed architecture is ready to accept new events.

\section{Evaluation}
\label{sec:ewaluacja}

\subsection*{Algorithm}

\begin{figure}[!t]
	\centering
	\includegraphics[width=3.45in]{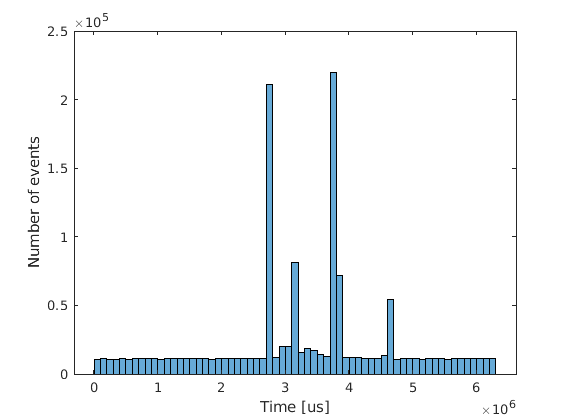}
	\caption{Histogram of events' timestamps.}
	\label{fig:recorded_hist}
\end{figure}

The described algorithm has been programmed and tested in the MATLAB computing environment.
Several datasets were used for the evaluation.
These sets were original and modified event camera recordings. 
The modification involved the addition of random artificial noise to the input data.
This made it easier to assess the effectiveness of the algorithm for data with various levels of noise. 
The labelling of the generated disturbances also allowed to assess what part of artificial and real events was removed by the proposed filter.
Filtration efficiency was tested on three different event camera recordings.
For each of the recordings used, an estimate of the minimum number of noisy events was made on the basis of a timestamp histogram. This allows one to assess how much of the original events should be removed by filtration.
The histogram for the second recording is presented in Figure \ref{fig:recorded_hist}.

The visible peaks and increased values correspond to objects moving in the field of view of the sensor. 
However, most of the histogram fields show mainly noise. 
The fields responsible for object movement have therefore been discarded, and the average value has been determined from the remaining elements. 
This value was assumed to be the average number of noise events per time unit. 
Then, this value was multiplied by the length of the entire recording. 
However, it should be noted that the recordings show an increased number of noises in areas for which there was activity shortly before (some object was moving). 
They are not included in the estimate. 
Therefore, the calculated number of noises is an estimate of the lower limit. 
In reality, it is almost certainly higher.


The filtering efficiency for different datasets is shown in Tables \ref{tab:80_balls_eval}, \ref{tab:recorded_eval}, and \ref{tab:person_eval}. 
It has been calculated for the basic versions of the recordings, as well as for those modified by adding artificial noise.
The values in the first column represent the average number of artificial noisy events generated for each millisecond of recording. 
The second column denotes the part of the generated noise that remains after filtering has been performed. 
The third column denotes the part of the original events that has not been filtered out. 
For each test case, the area size was \(16 \times 16\) and the update factor was \(0.25\). 
However, the length of the filter was changed.


The first set is a recording \textit{Falling 80 balls} prepared by \textit{Prophesee} and made available on their website: \url{https://docs.prophesee.ai/stable/datasets.html}. 
A \(640 \times 480\) resolution sensor was used for the recording. 
It contains a small amount of noise as the estimate performed indicates that only \(0.44\%\) of the events are disturbances. 
The filter length was \(200 \si{\micro s}\).

\begin{table}[!t]
	\centering
	\caption{Algorithm efficiency for the first dataset}
	\begin{tabular}{| c | c | c |}
		\hline 
		{Noises [events/ms]} & {Noise Remaining} & {Original Remaining} \\	
		\hline
		0       & --    & 95.91\% \\	
		\hline
		200     & 0.43\%    & 95.94\% \\
		\hline
		500     & 0.44\%    & 96.00\%  \\
		\hline
		1000    & 0.44\%    & 96.10\%  \\
		\hline
		2000    & 0.44\%    & 96.33\%  \\
		\hline
		5000    & 0.48\%    & 96.92\%  \\
		\hline
		10000   & 2.37\%    & 97.73\%  \\
		\hline
		20000   & 37.77\%   & 99.04\%  \\
		\hline
	\end{tabular}
	\label{tab:80_balls_eval}
\end{table}


The second set is a recording of several falling objects (screws). 
It was registered with Prophesee's \textit{EVK1 - Gen4 HD} evaluation kit, which includes a \(1280 \times 720\) resolution sensor. 
It contains a very high number of disturbances -- the estimation made indicates that even above \(54.03\%\) of the total number of events.
The filter length was \(1000 \si{\micro s}\).

\begin{table}[!t]
	\centering
	\caption{Algorithm efficiency for the second dataset}
	\begin{tabular}{| c | c | c |}
		\hline 
		{Noises [events/ms]} & {Noise Remaining} & {Original Remaining} \\	
		\hline
		0       & --        & 43.18\% \\	
		\hline
		200     & 0.05\%    & 43.22\% \\
		\hline
		500     & 0.05\%    & 43.27\%  \\
		\hline
		1000    & 0.06\%    & 43.36\%  \\
		\hline
		2000    & 0.06\%    & 43.59\%  \\
		\hline
		5000    & 1.10\%    & 44.99\%  \\
		\hline
		10000   & 23.19\%   & 58.69\%  \\
		\hline
		20000   & 86.04\%   & 92.66\%  \\
		\hline
	\end{tabular}
	\label{tab:recorded_eval}
\end{table}


The third set is a recording of a person walking in close proximity to the camera. 
It was done with the same evaluation set. 
The calculated estimate indicates that at least \(3.04\%\) of the events are noise. 
The filter length was \(2000 \si{\micro s}\).

\begin{table}[!t]
	\centering
	\caption{Algorithm efficiency for the third dataset}
	\begin{tabular}{| c | c | c |}
		\hline 
		{Noises [events/ms]} & {Noise Remaining} & {Original Remaining} \\	
		\hline
		0       & --        & 95.82\% \\	
		\hline
		200     & 3.86\%    & 95.92\% \\
		\hline
		500     & 3.95\%    & 96.04\%  \\
		\hline
		1000    & 4.24\%    & 96.24\%  \\
		\hline
		2000    & 6.34\%    & 96.67\%  \\
		\hline
		5000    & 36.65\%    & 98.22\%  \\
		\hline
		10000   & 89.56\%    & 99.75\%  \\
		\hline
		20000   & 99.94\%   & 100.00\%  \\
		\hline
	\end{tabular}
	\label{tab:person_eval}
\end{table}


When analysing the results presented, several things should be noted.
First, the filtering efficiency depends on the amount of noise. 
A larger number of noisy events per time unit increases the probability that several events occur in the same area in a small time interval. 
As a result, some of the disturbances may be classified as valid events.


Second, the length of the filter should be chosen based on the amount of noise at the sensor output.
It can be assumed that filters of lengths \(100 \si{\micro s}\), \(1000 \si{\micro s}\), and \(2000 \si{\micro s}\) stopped coping when the number of disturbances per millisecond was greater than 10000, 5000 and 2000, respectively.


The third thing is the fact that, choosing filtering parameters, apart from the amount of noise, one should also take into account the dynamics of the observed process (e.g. the speed of moving objects). 
Processes with low dynamics generate fewer events per time unit. 
Therefore, a filter of small length could remove a significant part of the relevant events. 
However, processes with high dynamics generate a large number of events in a short time, which makes it easier to distinguish them from disturbances.
Therefore, the filter parameters should be chosen taking into account both the amount of noise and the dynamics of the observed process.


Another conclusion concerns more unfiltered interference for the third set.
Even for the case with the lowest level of noise generated, the filter only removed above \(96\%\) of the noise.
In comparison, the values were above \(99\%\) in the others. 
This is because the object in the last set (a person passing close to the sensor) was significantly larger than the objects in the other cases. 
Therefore, the actual events were generated over a much larger area of the sensor, which caused more areas to pass events (also artificially generated noise). 
For large objects moving in the field of view of the camera, it is more difficult to perform effective filtering using the proposed approach.


A final observation based on the experiments performed concerns the removal of events for objects passing between areas. 
If an object moves in the camera's field of view, several events are deleted when this object passes between adjacent regions. 
The number of deleted events depends on the state of the filter and is consistent with the figure discussed previously \ref{fig:DiscardedEvents}. The deletion of events when passing between regions is the biggest drawback of the proposed filtering method (it has been significantly reduced by the global update but is still present).

\subsection*{Architecture}


The proposed architecture was described in the \textit{SystemVerilog} hardware description language. 
The tests of separate modules as well as of the whole system were carried out in the simulator of the \textit{Vivado 2020.2} IDE.
After confirming the full compatibility of the architecture simulation results with the results in the MATLAB computing environment, a test of the designed architecture was carried out on the \textit{Xilinx Zynq Ultrascale+ MPSoC} chip of \textit{Mercury+ XU9} module with the \textit{Mercury+ ST1} base card.


In order to perform an accurate comparison between the hardware results and the results in the \textit{MATLAB} environment, the test events were written to an additional BRAM memory, from which they were read during system operation and fed to the \texttt{EVENT IN} interface of the designed architecture. 
The output data and the contents of the BRAM memory in the \texttt{Time Map} module were in turn checked using the \textit{Integrated Logic Analyser} tool, which allows reading of the signal states in hardware without interrupting system operation. 
It was possible to achieve full compatibility of the results obtained in the hardware with the results obtained in the \textit{MATLAB}.


The implementation results of the discussed architecture were verified for a sensor resolution of \(640 \times 480\) and \(1280 \times 720\). In both cases, the area size was \(16 \times 16\).
The resource utilisation is shown in Table \ref{tab:zasoby}. It should be noted that even for high resolution the use of resources remains at very low level.
The maximum operating frequency for these implementations was \(387 \si{MHz}\) and \(361.5 \si{MHz}\) respectively. The maximum throughput also depends on the frequency of the global update. When performed every \(1 \si{ms}\), the throughputs are \(385.8 \si{MEPS}\) and \(357.9 \si{MEPS}\), respectively, while when performed every \(10 \si{ms}\), they are \(386.9 \si{MEPS}\) and \(361.1 \si{MEPS}\).
For the first case, Vivado's estimate of power consumption indicated \(0.899 \si{W}\), of which \(0.595 \si{W}\) was static power.
For the second case, the values were \(0.973 \si{W}\) and \(0.595 \si{W}\), respectively.


\begin{table}[!t]
	\centering
	\caption{Resource utilisation of the proposed architecture.}
	\begin{tabular}{| c | c | c |}
		\hline 
		{Resource} & {\(640 \times 480\)}   & \(1280 \times 720\)  \\	
		\hline
		LUT      & 2750 (1.19\%) & 7438 (3.23\%)\\	
		\hline
		LUTRAM   & 10 (0.01\%) & 10 (0.01\%)  \\
		\hline
		FF       & 1603 (0.35\%) & 4071 (0.88\%)\\
		\hline
		BRAM     & 2 (0.64\%) & 7.5 (2.40\%)   \\
		\hline
		BUFG     & 1 (0.18\%) & 1 (0.18\%)   \\
		\hline
	\end{tabular}
	\label{tab:zasoby}
\end{table}




\section{Conclusions}
\label{sec:podsumowanie}

In this paper, we have presented a new filtering algorithm for event sensors and its hardware architecture, which then we have implemented on the \textit{Xilinx Zynq Ultrascale+ MPSoC} chip of the \textit{Mercury+ XU9} module with the \textit{Mercury+ ST1} base card.
When designing the algorithm, we paid particular attention to reducing the memory used so that the architecture based on it could be implemented in low-memory chips even for high-resolution sensors.
While designing the hardware architecture, we paid special attention to maximising throughput, i.e. processing a new event in each clock cycle and achieving the highest possible clock frequency of the described architecture. 
As a result, we obtained a throughput of \(387 \si{MEPS}\) in normal operation mode and \(385.8 \si{MEPS}\) with a global update every \(1 \si{ms}\) for a resolution of \(640 \times 480\). For resolution \(1280 \times 720\), these values are \(361.5 \si{MEPS}\) and \(357.9 \si{MEPS}\), respectively.
It should also be noted that the designed architecture uses a very small fraction of the computational resources available in the used system (up to \(3.3\%\) for the LUT). 
The model of the algorithm proposed in the MATLAB computing environment can be found in the repository on the \textit{GitHub} platform: \url{https://github.com/vision-agh/DVS_FilterMatrixIIR}.

In Table \ref{tab:throughput} the throughput of the state-of-the-art filtering solutions was compared.
The computing platforms on which the results were achieved are also given. It should be noted that comparing solutions designed for different architectures is very difficult. This is due, for example, to the fact that on general-purpose processors it is relatively easy to use all the available computing power (at least for single cores). In contrast, the solution proposed in this paper uses only about 3\% of the available resources. Comparing implementations between different FPGAs is also difficult due to the different maximum operating frequencies of the computing elements. For this reason, the throughput of the solutions is simply given.
\begin{table}[!t]
	\caption{Comparison of filtering modules throughput}
	\begin{tabular}{| c | c | c |}
		\hline 
		{Resource} & {Platform} & {Throughput (MEPS)} \\	
		\hline
		This work & Xilinx Zynq Ultrascale+ MPSoC    & \(386.9\) \\	
		\hline
		\cite{bisulco2020near} & Spartan-6 FPGA     & \(50\)   \\
		\hline
		\cite{delbruck2008frame} & Pentium M processor   & \(10\) \\
		\hline
		\cite{linares2015usb3} & Lattice FPGA     & \(10\) \\
		\hline
		\cite{czech2016evaluating} & Intel Core i7-4790 & \(5.93\)  \\
		\hline
		\cite{mohamed2022dba} & Jetson TX2  & \(3\)  \\
		\hline
	\end{tabular}
	\label{tab:throughput}
\end{table}
To the best of our knowledge, the presented architecture provides the highest throughput of filtering data from dynamic vision sensors among the solutions described in the scientific literature.


The biggest drawback of the presented algorithm and the architecture based on it is the removal of several events when an object in the sensor's field of view passes between neighbouring areas. In future work, we will try to propose a mechanism that would be able to minimise or completely eliminate this effect. In addition, we plan to determine the effectiveness of our solution on a wider database, compare it with other methods for the same data and check the results in combination with other applications, such as object tracking.

\end{document}